\documentclass[10pt,twocolumn,letterpaper]{article}

\usepackage{iccv}
\usepackage{times}
\usepackage{epsfig}
\usepackage{graphicx}
\usepackage{amsmath}
\usepackage{amssymb}
\usepackage[linesnumbered,boxed]{algorithm2e}
\usepackage{multirow}
\include{pythonlisting}
\usepackage[title]{appendix}
\usepackage{stfloats}
\usepackage{booktabs,tabularx}
\makeatletter
\newcommand{\printfnsymbol}[1]{%
  \textsuperscript{\@fnsymbol{#1}}%
}

\usepackage[breaklinks=true,bookmarks=false]{hyperref}

\iccvfinalcopy 

\def\iccvPaperID{****} 

\begin{document}

\title{SynthText3D: Synthesizing Scene Text Images from 3D Virtual Worlds}

\author{\textsuperscript{1}Minghui Liao\thanks{Authors contribute equally.} , \textsuperscript{2}Boyu Song\printfnsymbol{1}, \textsuperscript{2}Shangbang Long\printfnsymbol{1}, \textsuperscript{1}Minghang He, \textsuperscript{3}Cong Yao, \textsuperscript{1}Xiang Bai\\
\textsuperscript{1}Huazhong University of Science and Technology, \textsuperscript{2}Peking University, \textsuperscript{3}Megvii Inc.\\
{\tt\small \{mhliao,minghanghe09,xbai\}@hust.edu.cn}\\ 
{\tt\small \{boyusong,shangbang.long\}@pku.edu.cn, yaocong2010@gmail.com}
}

\maketitle

\begin{abstract}
With the development of deep neural networks, the demand for a significant amount of annotated training data becomes the performance bottlenecks in many fields of research and applications. Image synthesis can generate annotated images automatically and freely, which gains increasing attention recently. In this paper, we propose to synthesize scene text images from the 3D virtual worlds, where the precise descriptions of scenes, editable illumination/visibility, and realistic physics are provided. Different from the previous methods which paste the rendered text on static 2D images, our method can render the 3D virtual scene and text instances as an entirety. In this way, real-world variations, including complex perspective transformations, various illuminations, and occlusions, can be realized in our synthesized scene text images. Moreover, the same text instances with various viewpoints can be produced by randomly moving and rotating the virtual camera, which acts as human eyes. The experiments on the standard scene text detection benchmarks using the generated synthetic data demonstrate the effectiveness and superiority of the proposed method. The code and synthetic data is available at: https://github.com/MhLiao/SynthText3D.
\end{abstract}
\section{Introduction}
Performances of deep learning methods are highly relevant to the quality and quantity of training data. However, it is expensive and time-consuming to annotate the training data in domain-specific tasks. Image synthesis provides an effective way to alleviate this problem by generating large amounts of data with detailed annotations for specific tasks, which is labor-free and immune to human errors.
As for scene text, it is elusive and prone to imprecision to annotate a training dataset that covers scene text instances in various occasions due to the diversity and variation of backgrounds, languages, fonts, lighting conditions, etc..
Therefore, synthetic data and related technologies are crucial for scene text tasks in research and applications. Moreover, in contrast to manually annotated data, whose labels are usually in line or word level, synthetic data can provide detailed information, such as character-level or even pixel-level annotations.

\begin{figure}[t]
    \centering
    \begin{tabular}{@{}c@{}}
        \includegraphics[width=1.0\linewidth]{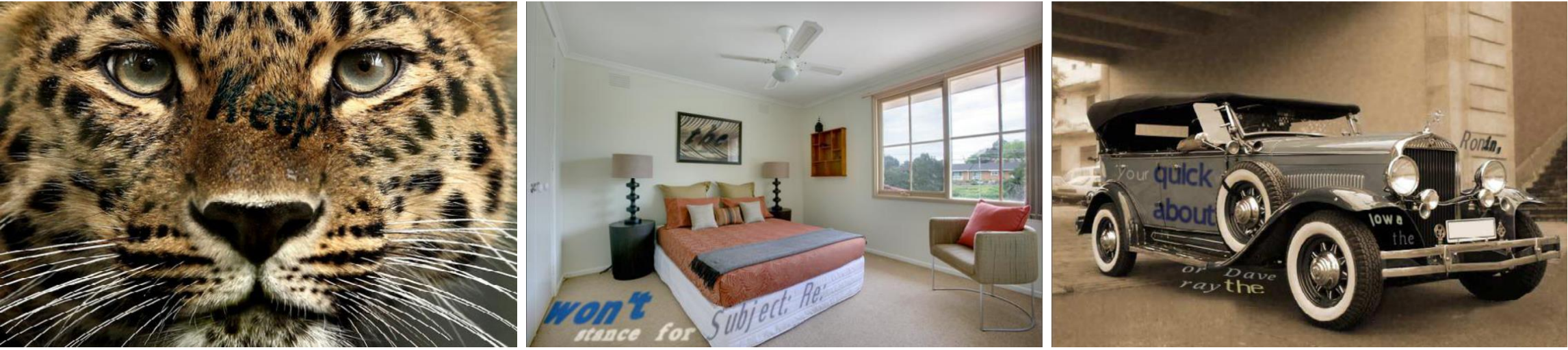}\\
        \small (a) Gupta et al.~\cite{gupta2016synthetic}
    \end{tabular}
    \hfill
    \begin{tabular}{@{}c@{}}
        \includegraphics[width=1.0\linewidth]{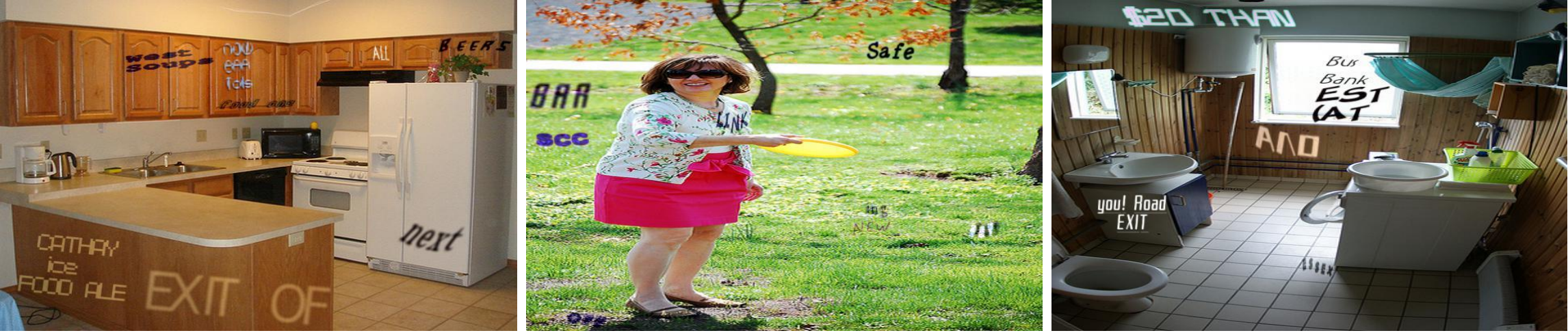}\\
        \small (b) Zhan et al.~\cite{zhan2018verisimilar}
    \end{tabular}
    \hfill
    \begin{tabular}{@{}c@{}}
        \includegraphics[width=1.0\linewidth]{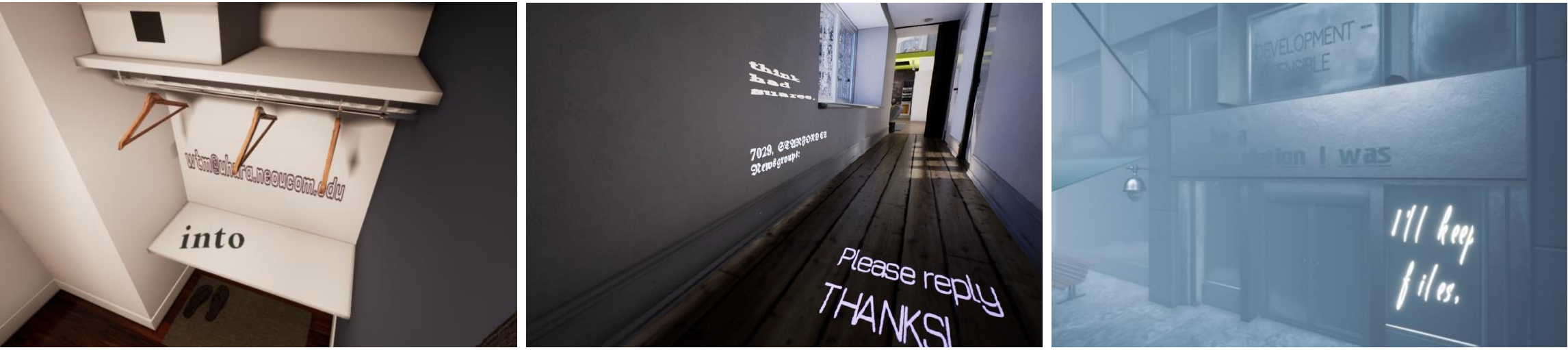}\\
        \small (c) Ours
    \end{tabular}
    \caption{Image samples synthesized by different methods. Compared to existing methods that render text with static background images, our method can realize realistic occlusions, perspective transformations and various illuminations.}
    \label{fig:introduction}
\end{figure}

Previous scene text image synthesis methods~\cite{jaderberg2014synthetic,gupta2016synthetic,zhan2018verisimilar} have shown the potential of the synthetic data in scene text detection and recognition. The essential target of the scene text synthesizers is to generate scene text images similar to real-world images. In the previous methods~\cite{gupta2016synthetic,zhan2018verisimilar}, static background images are first analyzed using semantic segmentation/gpb-UCM and depth estimation, then text regions are proposed according to the predicted depth and semantic information. Here, ``static" means single-view images, instead of multi-view images from the same scenarios.
Zhan et al.~\cite{zhan2018verisimilar} further improve visual similarity by introducing ``semantic coherence'', ``homogeneous regions'', and adaptive appearances.

These methods and their data are widely used in scene text detection and recognition, bringing promising performance gains. However, different from their synthesis procedures, humans usually perceive scene text from 3D worlds, where text may possess the variations of viewpoints, illumination conditions, and occlusions. However, such variations are difficult to produce by inserting text into static 2D background images due to the lack of 3D information. Moreover, the candidate text regions generated by the results of segmentation algorithms may be rough and imprecise. These factors lead to a gap between the text images synthesized by existing engines and real-world text images.

Instead, we propose an image synthesis engine incorporating 3D information, which is named as \emph{SynthText3D}, generating synthetic scene text images from 3D virtual worlds. 
Specifically, text instances in various fonts are firstly embedded into suitable positions in a 3D virtual world. Then we render the virtual scene containing text instances with various illumination conditions and different visibility degrees in the 3D virtual world, where text and scene are rendered as integrity. Finally, we set the camera with different locations and orientations to make the projected 2D text images in various viewpoints.

Benefiting from the variable 3D viewpoint and the powerful virtual engine, SynthText3D is intuitively superior to previous methods in visual effects (as shown in Fig.~\ref{fig:introduction}) because of the reasons below: 
1) Text and scenes in the 3D virtual world are rendered as an entirety, which makes the illumination/visibility, perspective transformation, and occlusion simulation more realistic.
2) We can obtain accurate surface normal information directly from the engine, which is beneficial to finding proper regions to place text instances.
3) SynthText3D can generate text instances with diverse viewpoints, various illuminations, and different visibility degrees, which is akin to the way of observing by human eyes.

The contributions of this paper are three-fold:
1) We propose a novel method for generating synthetic scene text images from 3D virtual worlds, which is entirely different from the previous methods which render text with static background images.
2) The synthetic images produced from 3D virtual worlds yield fantastic visual effects, including complex perspective transforms, various illuminations, and occlusions. 
3) We validate the superiority of the produced synthetic data by conducting experiments on standard scene text detection benchmarks. Our synthetic data is complementary to the existing synthetic data.

\section{Related Work}
\subsection{Image Synthesis in 3D Virtual Worlds}
Image synthesis with 2D images is popular in various tasks, such as human pose generation~\cite{zhu2019progressive}.
Recently, synthesizing images with 3D models using graphics engines have made a figure and attracted much attention in several fields, including human pose estimations~\cite{varol2017learning}, indoor scene understanding~\cite{papon2015semantic,mccormac2016scenenet}, outdoor/urbane scene understanding~\cite{ros2016synthia,saleh2018effective}, and object detection~\cite{peng2015learning,tremblay2018falling,hinterstoisser2019annotation,yangyangarpnet,cao2017learning,cao2016pedestrian}. 
The use of 3D models falls into one of the following categories: 
(1) Rendering 3D objects on top of static background real-world images~\cite{peng2015learning,varol2017learning}; 
(2) Randomly arranging scenes filled with objects~\cite{papon2015semantic,mccormac2016scenenet,ros2016synthia,hinterstoisser2019annotation};
(3) Using commercial game engine, such as Grand Theft Auto V (GTA V)~\cite{craig_quiter_2018,martinez2017beyond,saleh2018effective} and the UnrealCV Project~\cite{qiu2016unrealcv,ganoni2017framework,tremblay2018falling}.

\subsection{Synthetic Data for Scene Text}
There have been several efforts~\cite{wang2012end,jaderberg2014synthetic,gupta2016synthetic,zhan2018verisimilar,zhan2019spatial} devoted to the synthesis of images for scene text detection and recognition. 
Synth90K~\cite{jaderberg2014synthetic} proposes to generate synthesized images to train a scene text recognizer.
Text instances are rendered through several steps, including font rendering, bordering/shadowing, coloring and projective distortion, before they are blended with crops from natural images and mixed with noises.
However, the generated images are only cropped local regions which can not be directly used to train text detectors. 
Zhan et al.~\cite{zhan2019spatial} propose to use GAN~\cite{goodfellow2014generative} to implant style-consistent text into natural images, but they are only image crops and can not be used in text detection. 
SynthText~\cite{gupta2016synthetic} is then proposed to synthesize scene images with text printed on them for scene text detection.
It searches for suitable areas, determines orientation, and applies perspective distortion according to semantic segmentation maps and depth maps, which are predicted by existing algorithms.
The synthesized images have more realistic appearances as text instances seem to be correctly placed on the surface of objects in images.
To compensate for the drawback that SynthText would print text onto unreasonable surfaces such as human faces which contradicts real-world scenarios, Zhan et al.~\cite{zhan2018verisimilar} proposed selective semantic segmentation in order that word instances are only printed on sensible objects. 
They further replace the coloring scheme in SynthText and render the text instances adaptively to fit into the artistic styles of their backgrounds. 

Different from these methods, which render text with static images, our method renders the text and the scene as integrity in the 3D virtual worlds.

\subsection{Scene Text Detection}
Scene text detection has become an increasingly popular and essential research topic~\cite{ye2015text,bai2018deep}. 
There are mainly two branches of ideology in fashion: \textit{Top-down} methods~\cite{yuliang2017detecting,liao2017textboxes,ma2018arbitrary,Liu2017Deep,He_2017_ICCV,Zhou_2017_CVPR,liao2018rotation,liao2019mask} treat text instances as objects under the framework of general object detection~\cite{ren2015faster,liu2016ssd}, adapting those widely recognized object detection methods for unique challenges in scene text detection.
These include rotating bounding boxes~\cite{liao2018textboxes++,liao2018rotation} and direct regression~\cite{Zhou_2017_CVPR,He_2017_ICCV} for multi-oriented text, region proposals of varying sizes for long text. 
\textit{Bottom-up} methods~\cite{Shi_2017_CVPR,wu2017self,long2018textsnake,Deng2018,Lyu2018,liao2019mask,chen2019irregular}, on the other hand, predict shape-agnostic local attributes or text segments, and therefore can predict irregular text~\cite{long2018textsnake} or very long text instances~\cite{Lyu2018}. 
These deep-learning frameworks require a large amount of labeled data, making it promising for the development of synthesis algorithms for scene text images.

\begin{figure*}[ht]
\centering
\includegraphics[width=0.98\linewidth]{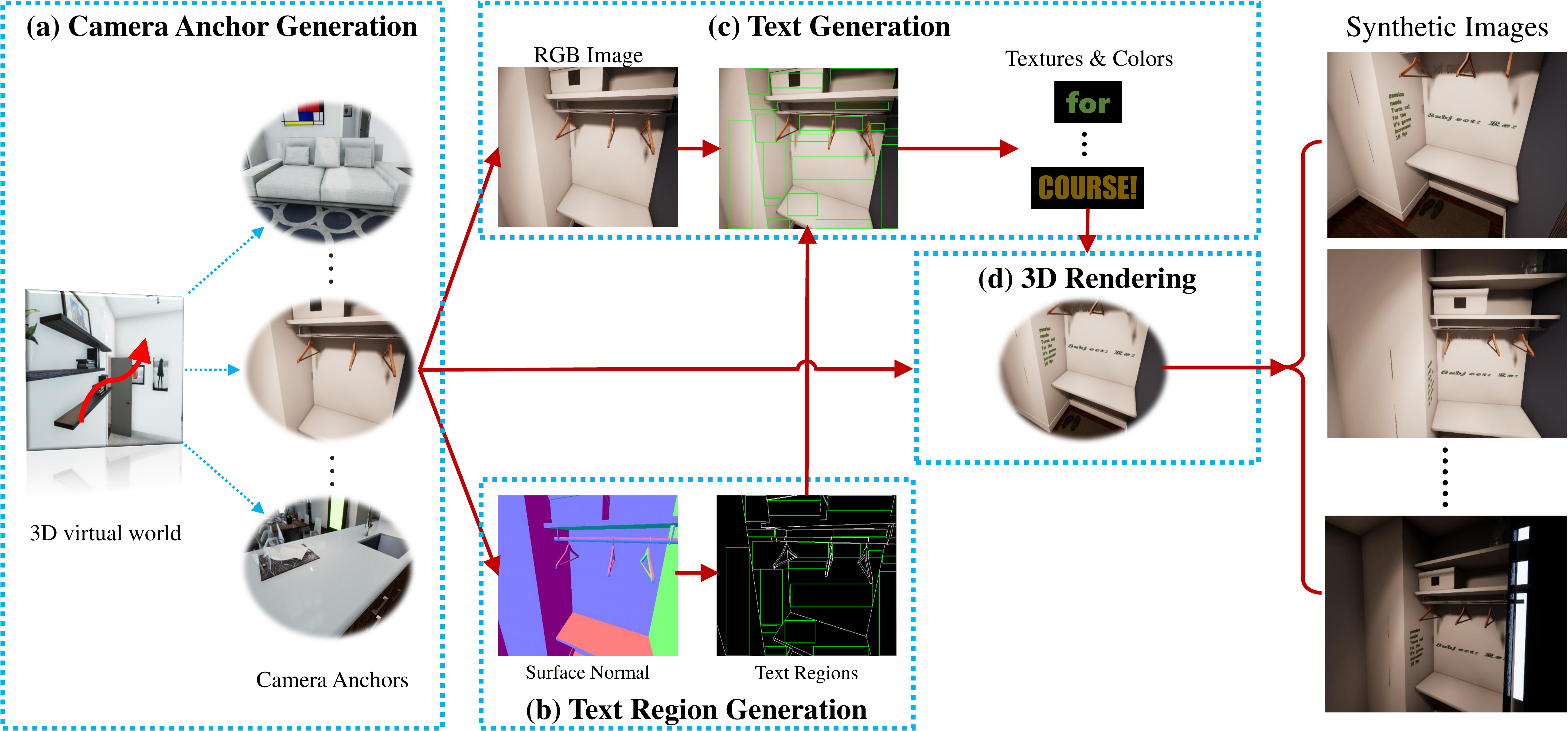}
\caption{The pipeline of SynthText3D. The blue dashed boxes represent the corresponding modules described in Sec.~\ref{sec:methodology}.}
\label{fig:pipeline}
\end{figure*}

\section{Methodology}\label{sec:methodology}
SynthText3D provides an effective and efficient way to generate large-scale synthetic scene text images. The 3D engine in our method is based on Unreal Engine 4 (UE4), with an UnrealCV~\cite{qiu2016unrealcv} plugin, which provides high rendering quality, realistic physics, editable illumination, and environment. 
We use $30$ 3D scene models including $17$ indoor models and $13$ outdoor models from the official marketplace\footnote{\url{https://www.unrealengine.com/marketplace}} of Unreal Engine to generate the synthetic data.

The pipeline of Synthtext3D mainly consists of a \emph{Camera Anchor Generation} module (Sec.~\ref{subsec:camera_anchor}), a \emph{Text Region Generation} module (Sec.~\ref{subsec:text_region}), a \emph{Text Generation} module (Sec.~\ref{subsec:text_generation}), and a \emph{3D Rendering} module (Sec.~\ref{subsec:3d_rendering}), as illustrated in Fig.~\ref{fig:pipeline}. 

First, we initialize a small number of camera anchors for each 3D model manually. Then, the RGB image and the accurate surface normal map of each camera anchor can be obtained from the 3D engine. Next, the available text regions are generated based on the surface normal map. After that, we randomly select several regions from all the available text regions and generate text of random fonts, text content, and writing structures based on the size of the text regions. Text colors of the selected text regions are generated according to the background RGB image of the corresponding regions. Finally, we map the 2D text regions into the 3D virtual world and place the corresponding text in it. Synthetic images with various illuminations and different viewpoints can be produced from the rendered 3D virtual world.

\subsection{Camera Anchor Generation} \label{subsec:camera_anchor}
Previous image synthesis methods for scene text conduct manual inspection~\cite{gupta2016synthetic}, or use annotated datasets~\cite{zhan2018verisimilar} to discard the background images which contain text.
Different from these methods which need a large number of static background images, we construct a small set of camera viewpoints (about 20 to 30 for each 3D model) in the virtual scenes, which are treated as the initial anchors. During collection, the operator controls the camera navigating in the scene, choosing views that suitable for placing text.

\begin{figure}[ht]
\centering
\includegraphics[width=0.9\linewidth]{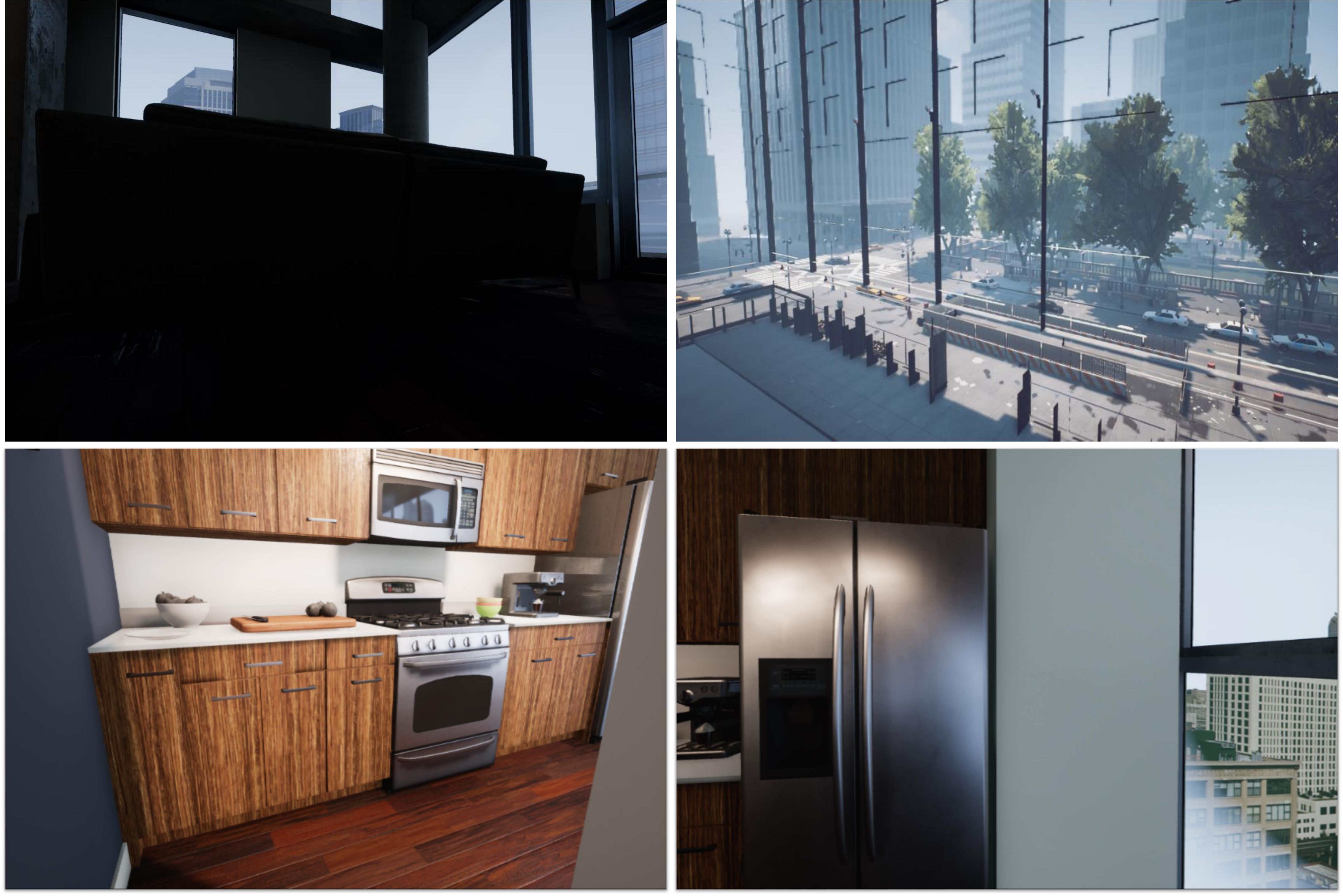}
\caption{Camera anchors. The first row depicts the randomly produced anchors (left: in dim light; right: inside a building model.) and the second row shows the manually selected anchors.}
\label{fig:bad_vp}
\end{figure}

We select the camera anchors following a simple rule: at least one suitable region exists in the view. The human-guided camera anchor generation can get rid of unreasonable anchors such as these insides of objects or in the dim light (see Fig.\ref{fig:bad_vp}).

\begin{figure*}[htbp]
    \centering
        \begin{tabular}{@{}c@{}}
            \includegraphics[width=0.18\textwidth]{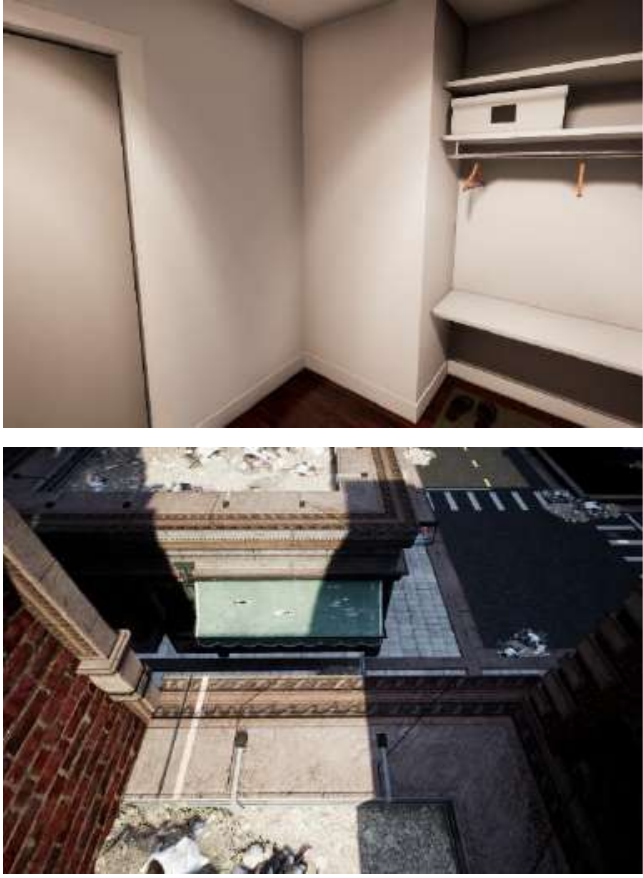}\\
            \small (a)
        \end{tabular}
        \begin{tabular}{@{}c@{}}
            \includegraphics[width=0.18\textwidth]{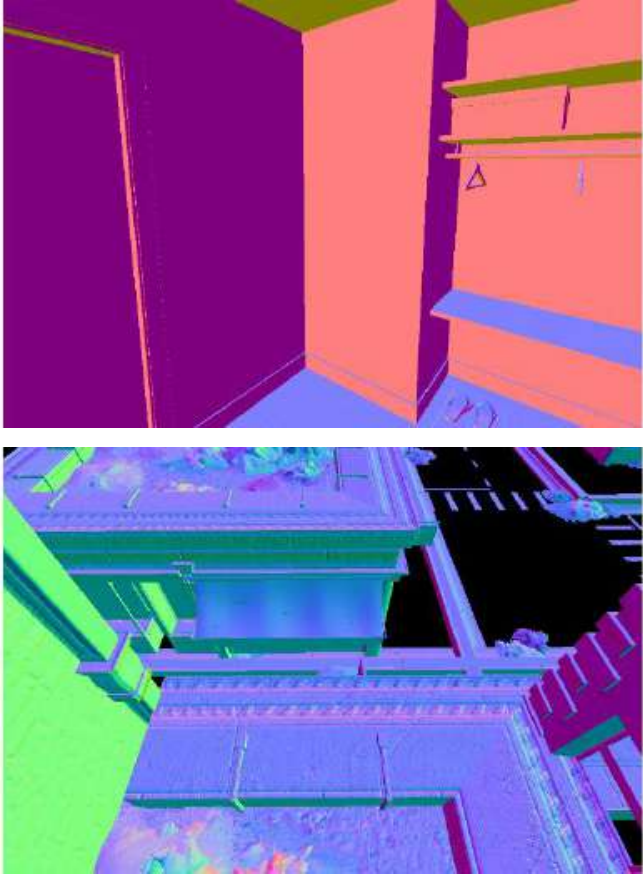}\\
            \small (b)
        \end{tabular}
        \begin{tabular}{@{}c@{}}
            \includegraphics[width=0.18\textwidth]{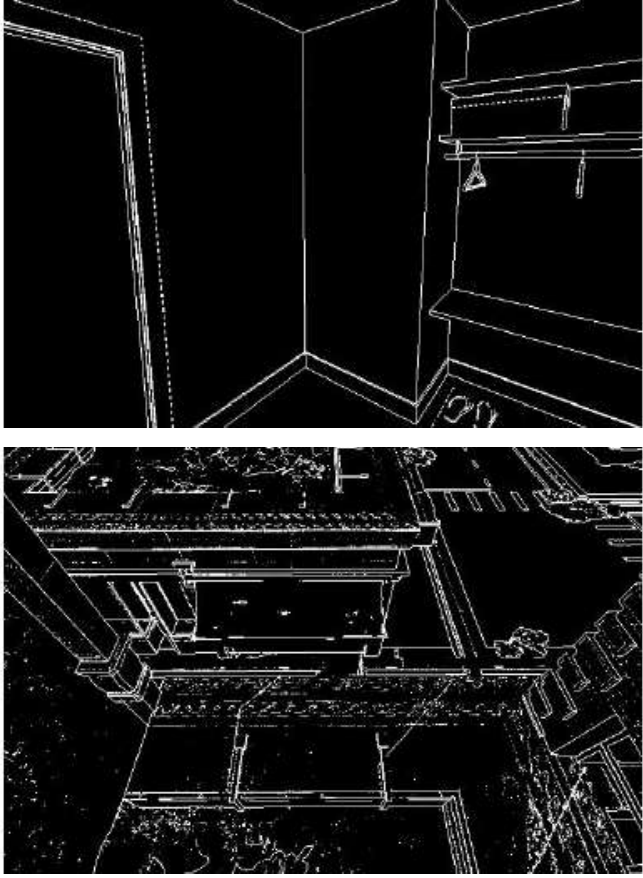}\\
            \small (c)
        \end{tabular}
        \begin{tabular}{@{}c@{}}
            \includegraphics[width=0.18\textwidth]{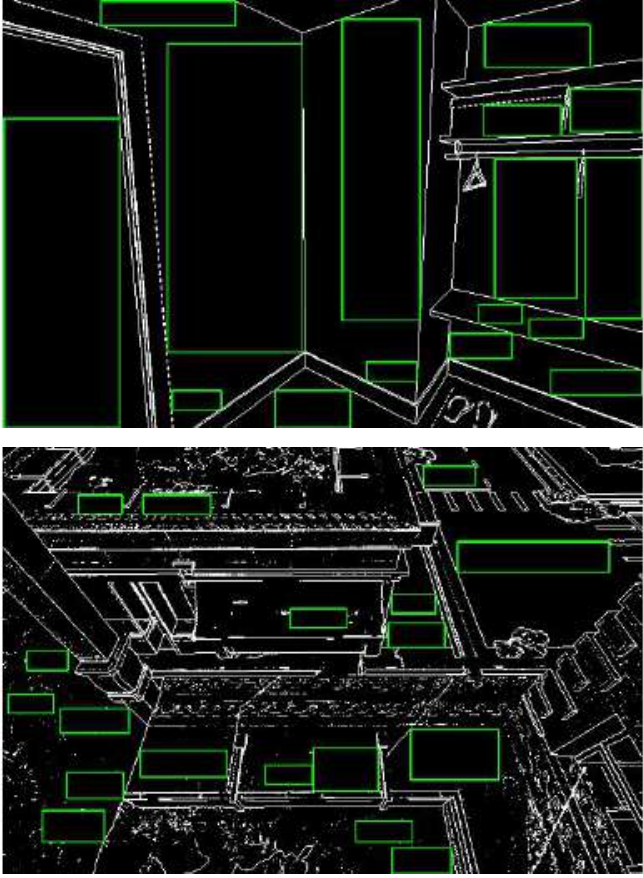}\\
            \small (d)
        \end{tabular}
        \begin{tabular}{@{}c@{}}
            \includegraphics[width=0.18\textwidth]{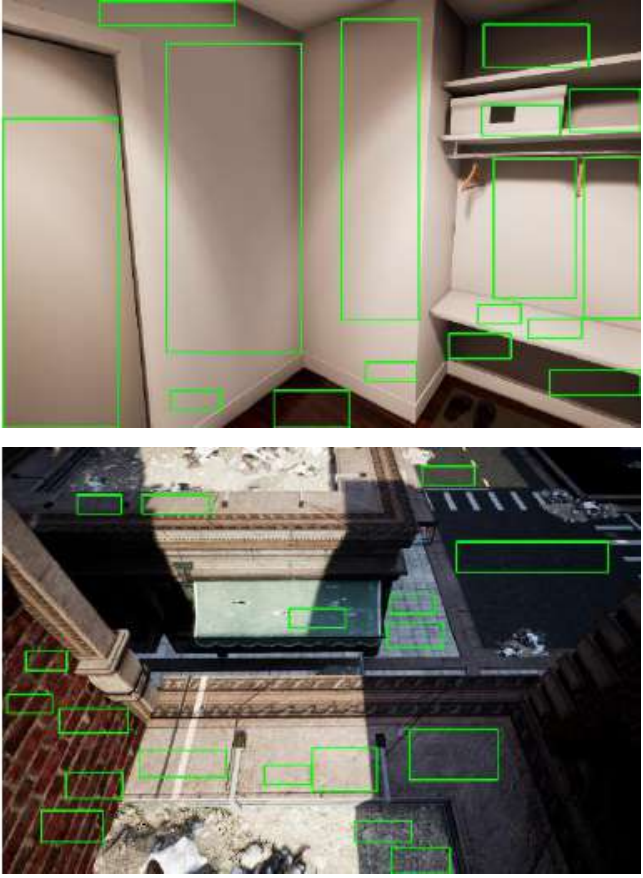}\\
            \small (e)
        \end{tabular}
    \caption{Illustration of \emph{Text Region Generation}. (a) are the original images; (b) are the surface normal maps; (c) are the normal boundary maps; (d) and (e) are the generated text regions.}
    \label{fig:normal_pipeline}
\end{figure*}

\subsection{Text Region Generation}\label{subsec:text_region}

Given a camera anchor, we can obtain a visible portion of the 3D scene, which contains the RGB image, the depth image, and the surface normal map of the view. Here, the generation of the text regions is based on the surface normal map.
A surface normal of a coordinate point in a 3D virtual world is defined as a unit vector that is perpendicular to the tangent plane to that surface at the current coordinate point.
Thus, the surface normal map is invariant to the viewpoint of the camera, which is more robust than the depth map.
First, we generate a normal boundary map according to the surface normal map. Then, a stochastic binary search algorithm is adopted to generate text regions.

\subsubsection{Normal Boundary Map}\label{subsubsec:normal_boundary}
Gupta et al.~\cite{gupta2016synthetic} obtain suitable text regions using gpb-UCM segmentation\cite{arbelaez2011contour} of images. Zhan et al.~\cite{zhan2018verisimilar} use saliency maps estimated by the model from \cite{lu2014robust} and ground truth semantic maps in \cite{lin2014microsoft} to extract suitable text embedding locations. However, the maps produced by extra models can be inaccurate and blurred (see Fig.~\ref{fig:depth_compare}) due to the gap between the training data and the testing data, especially for the boundary of objects or scenes, which tends to cause ambiguity for location extraction. Benefiting from the accurate representation of the 3D virtual engine, accurate depth maps, and normal maps can be obtained directly. Moreover, for real images, text instances are not strictly consistent with the appearance boundaries since they contain only 2D information. Normal boundaries contain 3D information that describes the smoothness of surface regions. Thus, normal boundaries are more suitable than appearance boundaries.

Intuitively, a region where all pixels share similar surface normal vectors is located inside the same object surface and is thus more suitable for text placement than a rugged one. Thus, a surface normal boundary map, where regions with drastically different normal vectors are separated, can be used to extract the text embedding region. We can generate a normal boundary map from a surface normal map using a simple transformation equation: 
\begin{equation}
  B_{i,j}= 
  \begin{cases}
    1& \max(\left | N_{i,j}-N_{i,j}^{0}\right |,\dots, 
\left | N_{i,j}-N_{i,j}^{k}\right |)>t, \\
    0& \text{otherwise}
  \end{cases}
\end{equation}
where $t$ is set as $100$; $N_{i,j}$ is a vector at position $(i, j)$ in the surface normal map $N$; $B_{i,j}$ is a value of position $(i, j)$ in the normal boundary map $B$; $\{N_{i,j}^{0},\dots N_{i,j}^{k} \}$ is a group of normal vectors of the 4 adjacent positions (except for the boundary position of the map). $\left|.\right|$ is the L1 norm function, $t$ denotes the threshold of L1 norm difference between a normal vector and its neighboring vector.
Examples of normal boundary map can be found in Fig.~\ref{fig:normal_pipeline} (c).

\begin{algorithm*}[ht]
\SetAlgoLined

\KwIn{binary normal boundary map: $M\in R^{H \times W}$, initial center point: $(X_0, Y_0)$}

\KwOut{top-left point:($X_1$, $Y_1$), bottom-right point: ($X_2$, $Y_2$)}

$S_W = 96$; $S_H = 64$; // \textit{The width and height of minimal initial box as mentioned in Sec.~\ref{subsubsec:anchor_search}.}

$X_{(1,upper)}=0$; $X_{(1, lower)}=X_0 - S_W / 2$\; 

$Y_{(1,upper)}=0$; $Y_{(1, lower)}=Y_0 - S_H / 2$\;

$X_{(2, lower)}=X_0 + S_W / 2$; $X_{(2,upper)}=W$\;

$Y_{(2, lower)}=Y_0 + S_H / 2$; $Y_{(2,upper)}= H$\;

\While{expandable} 
{

// \textit{``expandable" indicates if it was successfully expanded (not reaching boundaries) in the last step.}

randomly select $side$ from [$left$, $right$, $top$, $bottom$]\;

\uIf{$side$ is left $\And$ $abs(X_{(1, lower)}-X_{(1,upper)})>1$}
 {
      $mid= \lfloor(X_{(1, lower)}+X_{(1,upper)})/2\rfloor$\;
      
      \uIf{$sum(M[Y_{(1, lower)}:Y_{(2, lower)}, mid:X_{(2, lower)}]) >= 1$}
         {
              $X_{(1,upper)} = mid$\;
         }
         \Else{
         $X_{(1, lower)} = mid$
         }
 }
 \Else{
...;  // \textit{update $X_{(2,\bullet)}$, $Y_{(1,\bullet)}$, $Y_{(2,\bullet)}$ similarly when} \textit{side} is \textit{right}, \textit{top}, \textit{bottom}\;
}

$X_1 = X_{(1, lower)}$; $Y_1 = Y_{(1, lower)}$\;
$X_2 = X_{(2, lower)}$; $Y_2 = Y_{(2, lower)}$\;

}

\textbf{return} $(X_1, Y_1)$, $(X_2, Y_2)$ 

\caption{Stochastic binary search}
\label{algorithm:sbs}
\end{algorithm*}
 
\subsubsection{Stochastic Binary Search}\label{subsubsec:anchor_search}
To extract all the available text regions exhaustively, we set a grid of initial candidate regions over the image, inspired from anchor-based object detectors such as SSD~\cite{liu2016ssd} and faster-RCNN~\cite{ren2015faster}.
We iterate through the initial candidate regions. At each initial location, we start with an initial rectangular bounding box with a minimum size that is large enough to place text, which is set as $64\times 96$ pixels. The stride of the initial rectangular boxes is randomly chosen from $(12, 24, 36)$.
To retrieve the maximum areas, we propose a \textit{stochastic binary search method}. 
At each search step, we randomly expand one side of the rectangle.
The expansion follows the rule of a binary search: the \textit{lower bound} is set as the current edge location, the \textit{upper bound} is set as the corresponding edge of the image. 
If the expanded rectangle does not cross the normal boundary, then the lower bound is updated to the middle point; otherwise, the upper bound is updated to the middle point. 
The algorithm converges when the upper bound and lower bound of each edge equals each other.
After box generation at all anchors, we check each box one by one in a random order, and if any box overlaps with any other box, it is discarded.
Such randomness can enhance the diversity of boxes. Note that \textit{stochastic binary search method} is aimed to randomly find suitable text regions. Thus, we don't strictly find the maximum regions. 
Examples of results generated by the algorithm can be found in Fig.~\ref{fig:normal_pipeline} (d). Pseudocode is shown in Algorithm~\ref{algorithm:sbs}. 

\subsection{Text Generation}\label{subsec:text_generation}
Once the text regions are generated, we randomly sample several available text regions for text rendering.
Given a text region along with its RBG image, the text generation module is aimed to sample text content with certain appearances, including the fonts and text colors.
In order to compare with SynthText~\cite{gupta2016synthetic} fairly, we use the same text sources as \cite{gupta2016synthetic}, which is from the Newsgroup 20 dataset. The text content is randomly sampled from the text sources with three structures, including words, lines (up to 3), and paragraphs. 

We randomly sample the fonts from Google Fonts\footnote{\url{https://fonts.google.com/}}, which is used to generate the texture of the text. The text color is determined by the background of the text region, using the same color palette model in \cite{gupta2016synthetic}. 
The texture and the color of each region will be fed into the 3D Rendering module to perform 3D rendering.

\subsection{3D Rendering}\label{subsec:3d_rendering}
The 3D rendering includes two sub-modules. The first is the text placing module, which is aimed to place text into the 3D virtual worlds. The second is the rendering module, where the illumination and visibility adjustment, viewpoint transformation, occlusions can be performed.

\begin{figure*}[htbp]
    \centering
    \includegraphics[width=0.6\textwidth]{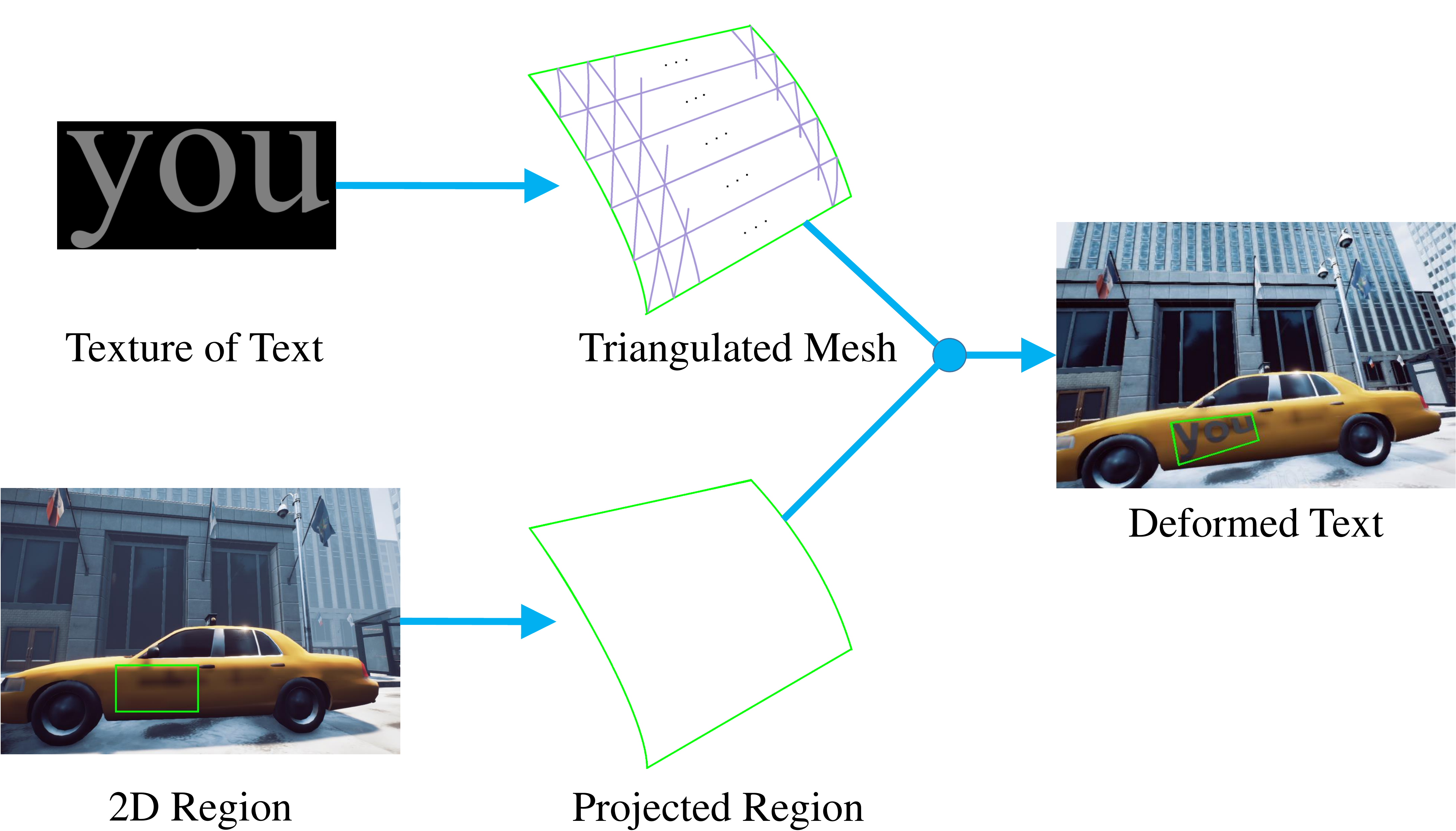}
    \caption{Illustration of text placing.}
    \label{fig:place_text}
\end{figure*}

\subsubsection{Text Placing}

With 2D text regions prepared, the text placing module first projects the 2D regions into the virtual scene by estimating the corresponding 3D regions. Then the text instances are deformed to fit the 3D surface properly.

\begin{figure}[htbp]
    \centering
    \includegraphics[width=0.4\textwidth]{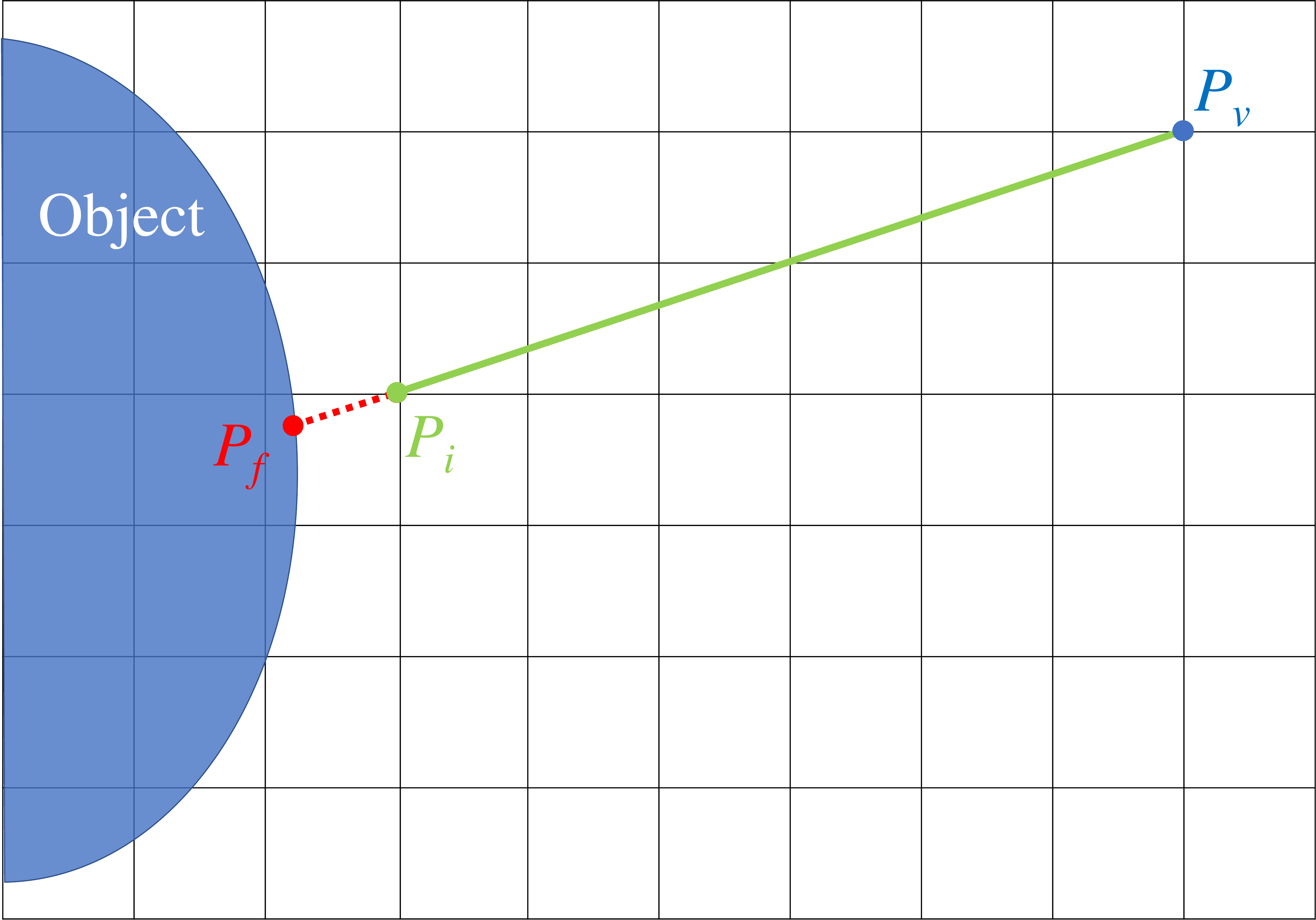}  
    \caption{The coordinates calculated by the integer depth and the float depth. $P_v$: coordinates of viewpoint; $P_i$: 3D location using integer depth; $P_f$:  3D location using floating point depth.}
    \label{fig:int_depth}
\end{figure}

\paragraph{2D-to-3D Region Projection} Previous works~\cite{gupta2016synthetic,zhan2018verisimilar} achieve this by predicting a dense depth map using CNN. In our method, we transform the 2D regions into the 3D virtual world according to the geometry information.

We adopt a coarse-to-fine strategy to project the 2D text regions into the 3D virtual worlds for projection precision. Assume that $p_j=(x_j, y_j)$ is a coordinate point in the 2D map, $d_j$ is the depth value of $p_j$, the coarse-grained 3D coordinates $P=(X_j, Y_j, Z_j)$ can be calculated as follows:
\begin{equation}
\label{function:camera_trans}
\left[ \begin{array}{ccc}
X_j\\
Y_j\\
Z_j
\end{array} 
\right ] = K \times d_j \times
\left[ \begin{array}{ccc}
x_j\\
y_j\\
1
\end{array} 
\right ]
\end{equation}
where $K$ is the internal reference matrix of the camera, which is an attribute parameter of the camera. 
The coarseness comes from the limitation of depth map where the pixel value is presented by an integer between [0,1024). However, as illustrated in Fig.\ref{fig:int_depth}, a deviation occurs between the location of integer depth and float one. 

In order to get the fine-grained coordinates, we adopt ray casting~\cite{Roth1982Ray}, a process of determining whether and where a ray intersects an object. As shown in Fig.\ref{fig:int_depth}, we initialize the ray $V=(P_i-P_v)$, where $P_i$ is a coarse point estimated in Eq.~\ref{function:camera_trans}. Then, the fine-grained point $P_f$ can be obtained by the ray casting, which detects the nearest hit point to $P_v$.

After projection, the regular bounding boxes are deformed to irregular 3D regions due to the geometric projective transformation shows in Eq.~\ref{function:camera_trans}. In the real world, the shapes of text are usually rectangles, and the orientation of the text instances are horizontal or vertical. We achieve this by clipping the sides of 3D quadrilaterals iteratively until it turns to 3D axis-aligned bounding boxes.

\paragraph{Text Deformation} In natural situations, not all text regions are flat planes, such as the surface of the bottle and the clothes, where text instaces need to be deformed to fit the target surface.

The deformation algorithm is illustrated as Fig.~\ref{fig:place_text}. We treat the text plane as a triangulated mesh, and text as the texture map of the mesh. Four corner vertexes of the mesh are first fixed to the region corners, and then intermediate vertexes are transformed to the nearest position on the surface of the target object. Finally, the texture coordinates of vertexes are estimated according to the Euclidean distance relative to the corner vertexes.

\begin{figure*}[ht]
    \centering
    \begin{tabular}{@{}c@{}}
        \includegraphics[width=1.0\linewidth]{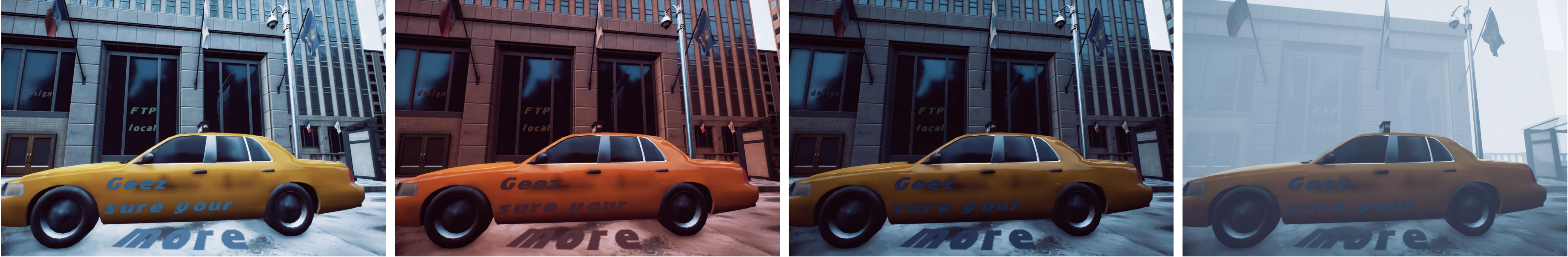}\\
        \small (a) Various illuminations and visibility of the same text instances.
    \end{tabular}%
    \hfill
    \begin{tabular}{@{}c@{}}
        \includegraphics[width=1.0\linewidth]{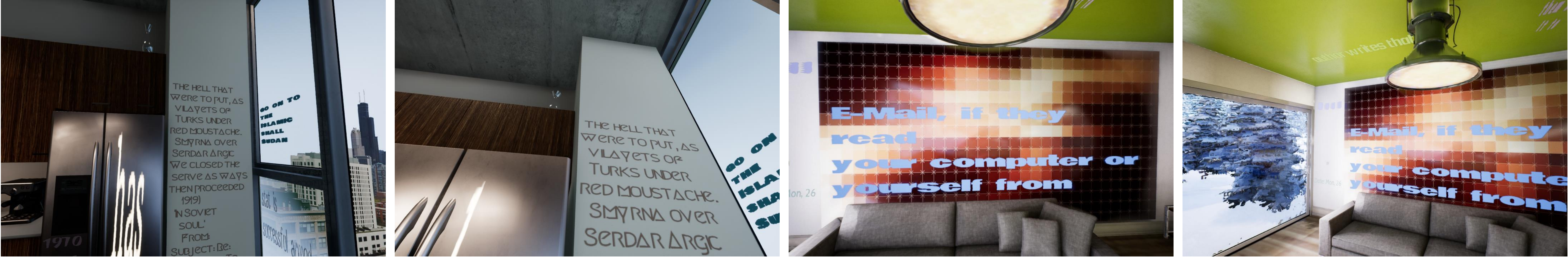}\\
        \small (b) Different viewpoints of the same text instances.
    \end{tabular}
    \hfill
    \begin{tabular}{@{}c@{}}
        \includegraphics[width=1.0\linewidth]{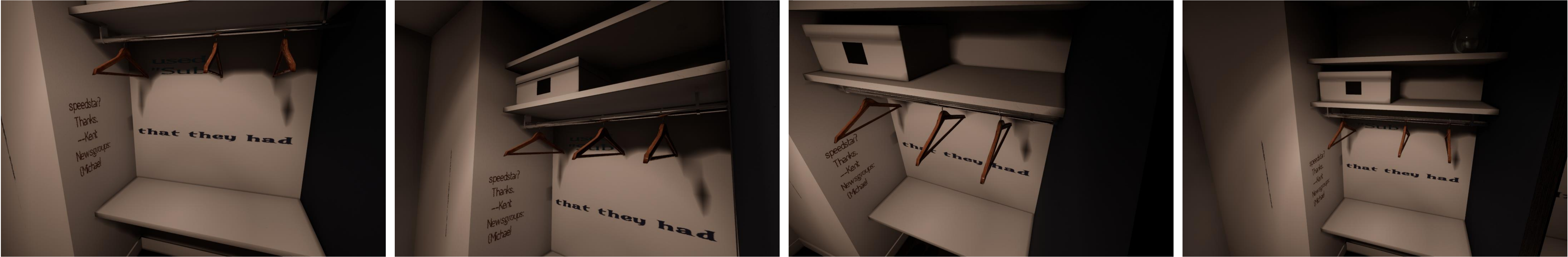}\\
        \small (c) Different occlusion cases of the same text instances.
    \end{tabular}
    \caption{Examples of our synthetic images. SynthText3D can achieve illuminations and visibility adjustment, perspective transformation, and occlusions.}
    \label{fig:examples}
\end{figure*}

\subsubsection{Rendering}
We establish several environment settings for each scene respectively. For each indoor scene, we build three kinds of illuminations: normal illuminance, brightness, and darkness. In addition to the illuminations, we add a fog environment for outdoor scenes.

After the text instances are placed into the 3D virtual world, we can augment to produce synthetic images with the following steps. First, we render the virtual objects along with text under various light intensity or color circumstances. Second, we randomly adjust the camera positions to conduct viewpoint transformation. For each camera anchor, we sample $20$ images using the rendering mentioned above. In this way, we can produce the synthetic images of the same text instances with various viewpoints, which contain rich perspective transformations.

\section{Experiments}
The synthetic images are synthesized with a workstation with a 3.20GHz Intel Core i7-8700 CPU, an Nvidia GeForce GTX 1060 GPU, and 16G RAM. The speed in the synthesizing period is about 2 seconds/image, with a resolution of $720 \times 1080$.

\subsection{Datasets}
\noindent\textbf{SynthText}~\cite{gupta2016synthetic} is a synthetic dataset consisting of 800k synthetic images, which are generated with 8k background images. Each background image is used to generate images of different text for 100 times. Other than using the whole set, we also randomly select 10k images from them, which is named as SynthText-10k.

\noindent\textbf{Verisimilar Image Synthesis Dataset} (VISD) is a synthetic dataset proposed in \cite{zhan2018verisimilar}. It contains 10k images synthesized with 10k background images. Thus, there are no repeated background images for this dataset. The rich background images make this dataset more diverse.

\noindent\textbf{ICDAR 2013 Focus Text} (ICDAR 2013)~\cite{karatzas2013icdar} is a dataset proposed in Challenge 2 of the ICDAR 2013 Robust Reading Competition. There are 229 images in the training set and 233 images in the test set.

\noindent\textbf{ICDAR 2015 Incidental Text} (ICDAR 2015)~\cite{karatzas2015icdar} is a popular benchmark for scene text detection. It consists of 1000 training images and 500 testing images, which are incidentally captured by Google glasses. 

\begin{figure*}[!t]
    \centering
    \begin{tabular}{@{}c@{}}
        \includegraphics[width=1.0\linewidth]{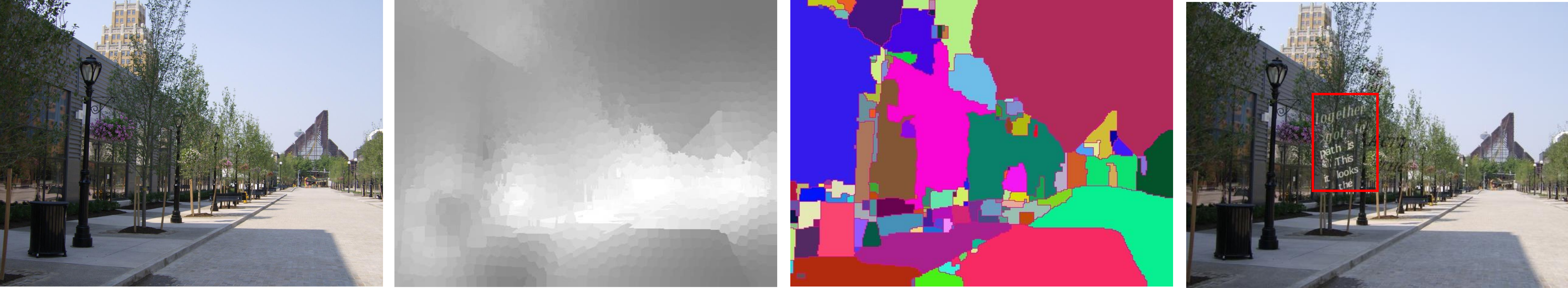}\\
        \small (a) SynthText
    \end{tabular}%
    \hfill
    \begin{tabular}{@{}c@{}}
        \includegraphics[width=1.0\linewidth]{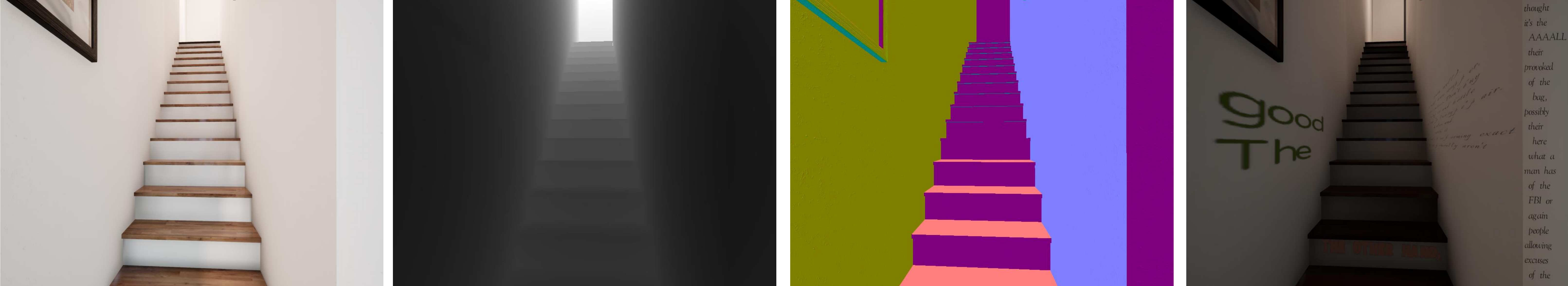}\\
        \small (b) Ours
    \end{tabular}
    \caption{Comparisons of the text regions between SynthText and ours. The first column: original images without text; the second column: depth maps; the third column: segmentation/surface normal maps; the last column: rendered images. The red boxes in the images mean unsuitable text regions.}
    \label{fig:depth_compare}
\end{figure*}

\noindent\textbf{MLT}\footnote{\url{http://rrc.cvc.uab.es/?ch=8&com=introduction}} is a dataset for ICDAR 2017 competition on multi-lingual scene text detection. It is composed of complete scene images that come from 9 languages representing 6 different scripts. There are 7,200 training images, 1,800 validation images and 9,000 testing images in this dataset. We only use the testing set to evaluate our models.

\subsection{Visual Analysis}
Two critical elements affect the visual effect of the synthetic data. One is the suitable text regions. Another one is the rendering effect.

\subsubsection{Suitable Text Regions}
Previous methods~\cite{gupta2016synthetic,zhan2018verisimilar} use saliency/gpb-UCM or depth maps to determine the suitable text regions in the static background. However, the segmentation or depth maps are estimated, which are not accurate enough. Thus, the text can be placed on unsuitable regions (see Fig.~\ref{fig:depth_compare}) (a). However, we can obtain accurate depth, segmentation, and surface normal information from the 3D engine easily. In this way, the text regions in our synthetic data are more realistic, as shown in Fig.~\ref{fig:depth_compare} (b).

\subsubsection{Rendering Effect}
Our SynthText3D can synthesis scene text images by rendering text and the background scene together. Thus, it can achieve more complex rendering than previous methods that render text with static images, such as the illumination and visibility adjustment, diverse viewpoints, occlusions, as shown in Fig.~\ref{fig:examples}. 

In Fig.~\ref{fig:examples} (a), we can see that the illuminations are diverse among the four images and the last image is in a foggy environment. 
The illumination effect can also be embodied in Fig.~\ref{fig:examples} (c), where the text and scene are in a dark environment. 
The viewpoint transformation is illustrated in Fig.~\ref{fig:examples} (b). 
Benefiting from the 3D perspective, we can change the viewpoint of the same text block to achieve realistic perspective transformation flexibly. Last but not least, occlusions can be performed by adjusting the viewpoint, as shown in Fig.~\ref{fig:examples} (c). 

\begin{table*}[ht]
\centering
\begin{tabularx}{1.0\linewidth}{@{}l*{9}X@{}}
\toprule
\multirow{2}{*}{Training data} & \multicolumn{3}{c}{ICDAR 2015} & \multicolumn{3}{c}{ICDAR 2013} & \multicolumn{3}{c}{MLT} \\ \cline{2-10} 
                               & P         & R        & F        & P         & R        & F        & P      & R      & F      \\ 
\midrule
SynthText 10k                  & 40.1      & 54.8     & 46.3     & 54.5      & 69.4     & 61.1     & 34.3   & 41.4   & 37.5   \\ 
SynthText 800k                 & 67.1      & 51.0     & 57.9     & 68.9      & 66.4     & 67.7     & 53.9   & 36.5   & 43.5   \\ 
VISD 10k                & 73.3      & 59.5     & 65.7     & 73.2      & 68.5     & 70.8     & 58.9   & 40.0   & 47.6   \\ 
Ours 10k (10 scenes)                       & 64.5      & 56.7     & 60.3   & 75.8      & 65.6     & 70.4    & 50.4   & 39.0   & 44.0   \\ 
Ours 10k (20 scenes)                      & 69.8      & 58.1     & 63.4   & 76.6      & 66.0     & 70.9    & 51.3   & 41.1   & 45.6   \\ 
Ours 10k (30 scenes)                      & 71.2      & 62.1     & \textbf{66.3}   & 77.1      & 67.3     & \textbf{71.9}    & 55.4   & 43.3   & \textbf{48.6}   \\ 

Ours 5k (10 scenes) + VISD 5k   &  71.1         & 64.4         & \textbf{67.6}         & 76.5          & 71.4         &  \textbf{73.8}        &  57.6      &  44.2      & \textbf{49.8}       \\ 
\bottomrule                     
\end{tabularx}
\caption{Detection results with different synthetic data. ``5k", ``10k'' and ``800k" indicate the number of images used for training. ``P'', ``R'', and ``F'' indicate precision, recall, and F-measure respectively, where $F=\frac{2 \times P \times R}{P + R}$. }
\label{tab:synth}
\end{table*}

As known, the rendering mentioned above is commonly seen in real-world images. However, limited to the static background images, which means the camera position, direction, and environment illumination are fixed, it is theoretically limited for previous methods to simulate such rendering. Thus, the rendering effect of our proposed synthetic data is more realistic.

\subsection{Scene Text Detection}
We conduct experiments on standard scene text detection benchmarks using an existing scene text detection model EAST~\cite{Zhou_2017_CVPR}. EAST is a previous state-of-the-art method that is both fast and accurate.
We adopt an open-source implementation\footnote{\url{https://github.com/argman/EAST}} to conduct all the experiments. We use ResNet-50~\cite{He_2017_Res} as the backbone. All models are trained with 4 GPUs with a batch size of 56.

\begin{figure*}[ht]
    \centering
    \includegraphics[width=1.0\textwidth]{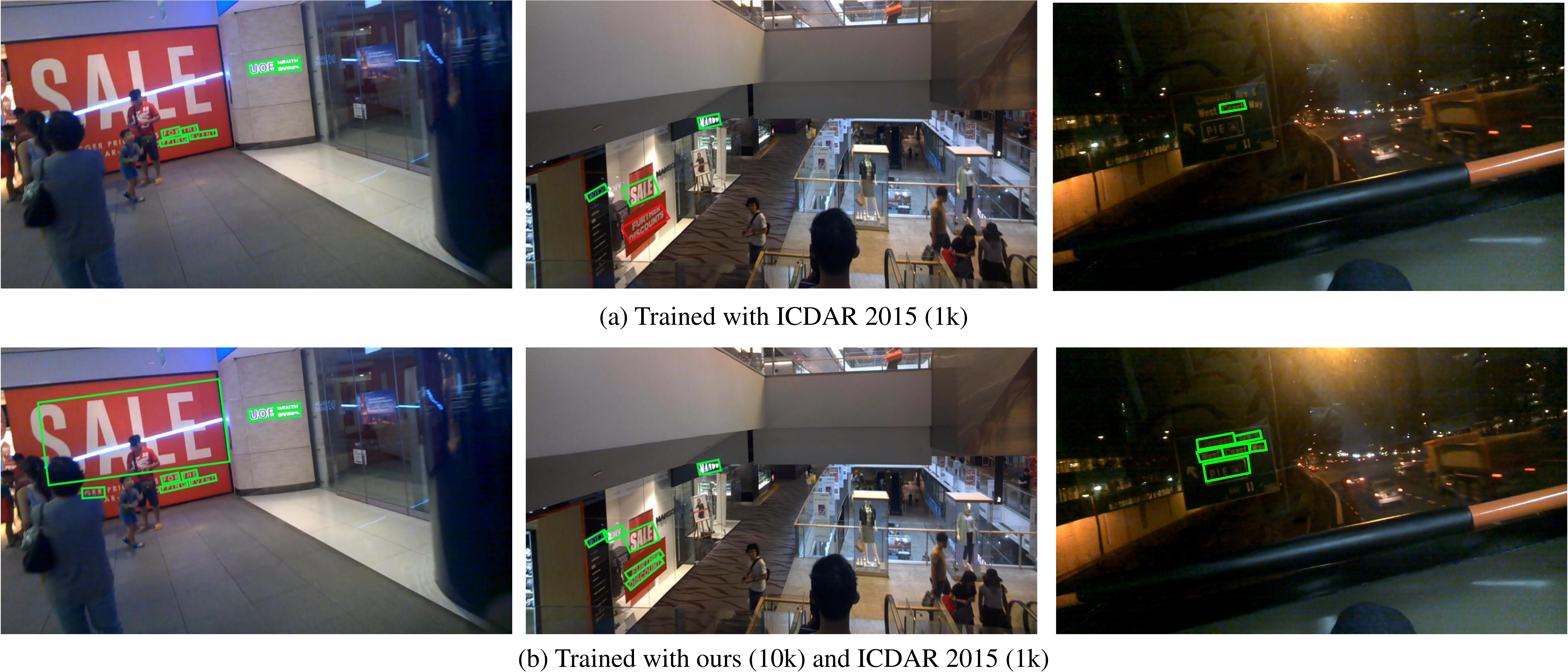}
    \caption{Detection results with different training set. Green boxes: correct results; Red boxes: wrong result. As shown, the occlusion and perspective cases can be improved when our synthetic data are used.}
    \label{fig:complementary}
\end{figure*}

\subsubsection{Pure Synthetic Data}
We train the EAST models with synthetic data to verify the quality of our synthetic images. The experimental results are listed in Tab.~\ref{tab:synth}. When using 10k images from 30 scenes, our synthetic data outperforms SynthText
by $20.0\%$, $10.8\%$ and $11.1\%$ in terms of F-measure on ICDAR 2015, ICDAR 2013 and MLT.

Note that ``Ours 10k'' surpasses ``SynthText 800k'' among all benchmarks, which effectively demonstrates the quality of our method, as the models trained with our synthetic data, compared to those trained with SynthText, can achieve better performance even with less training images.

It is unfair to directly compare with VISD as we use different text generation techniques.
The comparison with the milestone work SynthText is more convincing because we follow the same text generation method, including the corpus, color palette, et al. 
Nonetheless, with $30$ environments, our synthetic images achieve higher performance than the other two synthetic datasets on all test datasets. Specifically, on ICDAR 2013, $10$ scenes are enough to produce a comparable result. 

\subsubsection{Complementary Data} We conduct an experiment by training with mixed data (Ours 5K from 10 scenes + VISD 5k), achieving top accuracies as shown in Tab.~\ref{tab:synth} even though we do not use all the $30$ virtual environments. 
This phenomenon demonstrates that text generated from 3D is complementary to existing synthesized data. 
Specifically, while ours can simulate realistic real-world variations, synthetic data using natural images can help reduce the domain gap between real-world images and images from the virtual world.

\subsubsection{Scene Diversity} We also study how the number of virtual environments affects the potential of synthetic images. 
We generate 10k datasets from $10$, $20$, and $30$ scenes respectively. 
The results are also shown in Tab.\ref{tab:synth}. It is obvious that increasing the number of environments can improve performance significantly. As there are already many available virtual environments, we suppose that the potential of synthetic data from the 3D world is not yet fully exploited. 

\subsubsection{Combination of Synthetic Data and Real Data}
We conduct experiments with the combination of the synthetic data and the real-world data, to demonstrate that our synthetic can further improve the performance based on real-world data. The combination strategy is that we first pre-train the model with the synthetic data, and then fine-tune the model with real-world data. As shown in Tab.~\ref{tab:combination}, $1.4\%$ performance gain is acquired on ICDAR 2015 since some challenges cases are improved, such as occlusion and perspective transformation, as shown in Fig.~\ref{fig:complementary}. 

Note that all synthetic data listed in Tab.~\ref{tab:combination} can improve the performance similarly. The improvements are not significant since the synthetic data is only used for pre-training and the performance of the baseline is not weak.
However, the improvements are free of labor since the synthetic data can be automatically produced.

\begin{table}[ht]
\centering
\begin{tabularx}{1.0\linewidth}{@{}l*{3}X@{}}
\toprule
Training Data               & P    & R    & F   \\ 
\midrule
Real (1k)                   & 84.8 & 78.1 & 81.3 \\ 
SynthText (10k) + Real (1k) & 85.7 & 79.5 & 82.5 \\ 
VISD (10k) + Real (1k) & 86.5 & 80.0 & 83.1 \\ 
Ours (10k, 30 scenes) + Real (1k)   & \textbf{87.3} & \textbf{80.5} & \textbf{83.8} \\ 
\bottomrule
\end{tabularx}
\caption{Detection results on ICDAR 2015. ``Real'' means the training set of ICDAR 2015. ``VISD" is short for Verisimilar Image Synthesis Dataset~\cite{zhan2018verisimilar}.  ``P'', ``R'', and ``F'' indicate precision, recall, and F-measure respectively.}
\label{tab:combination}
\end{table}

\subsection{Limitations}
The main limitation of our method is that we need to select the camera anchors manually, though it is relatively easy to accomplish since there are only about 20-30 camera anchors for each virtual world. However, considering that previous methods also need manual inspection or annotations to filter background images that contain text, our manual selection is acceptable. We will try to improve it in the future work by introducing some heuristic rules or trained models to evaluate the generated camera anchors automatically.

\section{Conclusion}
In this paper, we proposed SynthText3D, a data synthesis engine, which can produce realistic scene text images for scene text detection from the 3D virtual worlds. Different from the previous methods which render text with static background images, SynthText3D has two advantages. One is that we can obtain accurate semantic and geometric attribute maps instead of the estimated segmentation or depth map, to determine the suitable text regions. Another one is that the text and scene can be rendered as an entirety in our method. In this way, we can achieve various illuminations and visibility, diverse viewpoints and occlusions. The visual effect and the experimental results on standard scene text benchmarks demonstrate the effectiveness of the synthesized data.

\section{Acknowledgements}
This work was supported by National Natural Science Foundation of China (61733007), to Dr. Xiang Bai by the National Program for Support of Top-notch Young Professionals and the Program for HUST Academic Frontier Youth Team 2017QYTD08.

{\small
\bibliographystyle{ieee}
\bibliography{reference}
}

\end{document}